\documentclass[letterpaper, 10 pt, journal, twoside]{IEEEtran}
\usepackage{amsmath,amsfonts}
\usepackage{algorithmic}
\usepackage{algorithm}
\usepackage{array}
\usepackage[caption=false,font=normalsize,labelfont=sf,textfont=sf]{subfig}
\usepackage{textcomp}
\usepackage{stfloats}
\usepackage{url}
\usepackage{verbatim}
\usepackage{graphicx}
\usepackage{cite}
\usepackage{multirow}    
\usepackage{makecell}    
\usepackage{booktabs}    
\usepackage[table]{xcolor}
\begin{document}

\title{TransNav: Transformer-Based End-to-End Navigation Method for Static and Dynamic Quadrotor Obstacle Avoidance}

\author{Jianghan Zhu\textsuperscript{1}, Zhuo Li\textsuperscript{1}, Yanjun Zhang\textsuperscript{1}, and Jian Sun\textsuperscript{1}, ~\IEEEmembership{Senior Member,~IEEE} 
\thanks{Manuscript received: April 22, 2026; Revised: July 24, 2026; Accepted: August 27, 2026. This article was recommended for publication by Associate Editor A. Candela and Editor S.-J. Chung upon evaluation of the reviewers’ comments. This work was supported in part by the National Natural Science Foundation of China under Grants 62495090, 62495095, 62688201, 62573050, and 62322304. \textit{(Corresponding author: Jian Sun.)}}
\thanks{\textsuperscript{1}The authors are with the State Key Laboratory of Autonomous Intelligent Unmanned Systems, School of Automation, Beijing Institute of Technology, Beijing 100081, China (e-mail: jianghan\_zhu@bit.edu.cn, zhuoli@bit.edu.cn, yanjun@bit.edu.cn, sunjian@bit.edu.cn). The source code is available at: https://github.com/jhzhubit/TransNav.git}
\thanks{Digital Object Identifier (DOI): 10.1109/LRA.2026.3734893}
}

\markboth{IEEE ROBOTICS AND AUTOMATION LETTERS. VOL. 11, NO. 11, NOVEMBER 2026}%
{Zhu \MakeLowercase{et al.}: TransNav: Transformer-Based End-to-End Navigation Method}


\maketitle

\begin{abstract}
Obstacle avoidance for autonomous quadrotors in cluttered environments remains challenging, especially when both dense static obstacles and moving obstacles are present. Existing methods are often tailored to either static or dynamic settings, which always limit their adaptability across diverse environments. This paper presents TransNav, a transformer-based end-to-end navigation method for quadrotor obstacle avoidance in both static and dynamic environments. TransNav employs a patch-based point-cloud encoding scheme that combines local feature extraction with global contextual modeling, enabling compact yet informative environment representation. The method integrates perception and navigation in a single architecture and directly predicts a finite-horizon action sequence, from which the first action is executed. Moreover, we develop an unsupervised learning strategy by incorporating a lightweight point-mass dynamics model, which directly optimizes trajectory-level costs over the predicted sequence through numerical gradient propagation. Thus, the policy can be efficiently trained without expert demonstrations. Simulation and real-world experiments show that the proposed method achieves strong performance in dense static environments and competitive obstacle-avoidance capability in dynamic scenarios, while maintaining low inference cost for onboard deployment.
\end{abstract}


\section{Introduction}
\IEEEPARstart{A}{utonomous} drones are widely deployed in scenarios such as exploration \cite{2021FUEL}, search and rescue \cite{alsamhi2022uav}, and inspection \cite{xu2023vision}. These environments are filled with densely distributed small obstacles and randomly moving dynamic obstacles, which require drones to perceive their surroundings in real time and generate immediate control commands to navigate safely. Traditional methods typically rely on manually designed perception-decision-control pipelines, where computational delays and cumulative errors severely degrade performance. In contrast, learning-based end-to-end navigation methods directly map raw sensory input to control commands, enabling high-performance real-time flight \cite{yu2024mavrl, song2023learning, lu2024you, hu2025seeing, zhang2026high, xu2025flying, xu2025flow, fan2025flying}. Recent studies using depth images \cite{lu2024you}, optical flow \cite{hu2025seeing}, or 3D LiDAR point clouds \cite{xu2025flying} as perception data have demonstrated outstanding results in dense environments.

Achieving a good balance between static-scene navigation and dynamic-obstacle avoidance remains difficult. Approaches designed for static environments \cite{lu2024you}, \cite{xu2025flying} react insufficiently to rapidly moving obstacles, while a few methods \cite{xu2025flow} tailored for dynamic scenarios tend to overfit to motion features, leading to degraded performance in dense static environments. Moreover, training strategies of the navigation policies in \cite{xu2025flow}, \cite{fan2025flying} rely heavily on training settings dominated by dynamic obstacles, which increase computational burden and fail to cover the full diversity of real-world obstacle motions, resulting in limited generalization. These limitations make it difficult to deploy an end-to-end policy reliably across a wide range of realistic conditions.

\begin{figure}[!t]
      \centering
      \includegraphics[width=0.95\columnwidth]{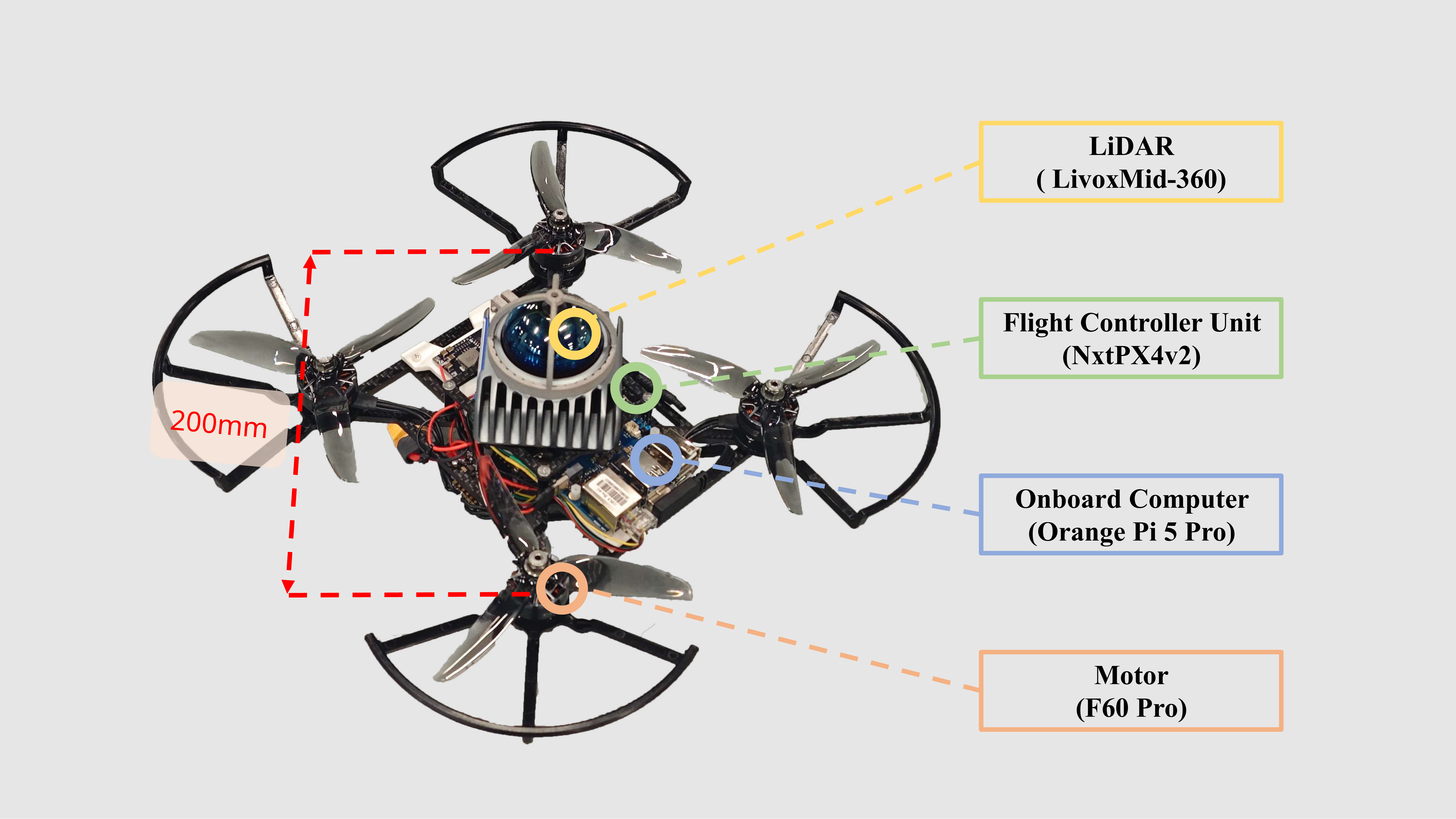}
      \caption{Hardware platform of the quadrotor in real-world experiments.}
      \label{fig:uav}
\end{figure}

To address these limitations, we propose TransNav, a transformer-based end-to-end navigation method for quadrotors using LiDAR point clouds. TransNav is designed to provide a more general policy for obstacle avoidance across both settings. On the one hand, to overcome the insufficient response speed of existing static-scene methods to moving obstacles, TransNav predicts a finite-horizon action sequence at a high frequency, achieving more effective reactions to obstacle motion. On the other hand, to ensure navigation performance in dense static environments, TransNav employs a patch-based LiDAR representation that preserves local geometric details while allowing global spatial context to be modeled compactly. Specifically, raw point clouds are first converted into a high-resolution distance map and then partitioned into patches inspired by vision transformers. A lightweight CNN and self-attention mechanism are further used to jointly encode local obstacle structure and longer-range spatial dependencies. In addition, an efficient unsupervised training strategy is introduced to directly optimize trajectory-level costs over a finite prediction horizon. By integrating a lightweight point-mass dynamics model into the learning pipeline, gradients of the full action sequence can be backpropagated directly without expert demonstrations or critic networks. This design enables fast convergence while maintaining strong sim-to-real generalization. The quadrotor used in real-world environments is shown in Fig. \ref{fig:uav}.

The main contributions are summarized as follows: 
\begin{itemize}
    \item 
    A transformer-based end-to-end navigation architecture is proposed for quadrotors, which fuses patch-based LiDAR distance-map representations with sampled reference-trajectory states to predict a finite-horizon action sequence within a unified framework.
\end{itemize}
\begin{itemize}
    \item 
    A lightweight perception-to-action pipeline is developed, which predicts action sequences for high-frequency receding-horizon execution. It improves the response ability of the policy to moving obstacles while maintaining strong navigation performance in dense static environments.
\end{itemize}
\begin{itemize}
    \item 
    {An efficient unsupervised training strategy for the proposed action-sequence policy is developed, which incorporates a lightweight point-mass dynamics model to directly optimize trajectory-level costs over the predicted action sequence without expert action labels.}
\end{itemize}

\section{Related work}

\subsection{Static Environment Navigation}
In static cluttered environments, most methods rely on optimization-based navigation pipelines that decompose the problem into perception, planning, and control. For example, Zhou et al. \cite{zhou2020ego} integrate grid maps with an optimization-based planning algorithm to enable real-time trajectory generation using onboard sensors such as LiDAR or depth cameras. However, high-resolution grid maps increase computational burden, while low-resolution maps fail to effectively preserve perception of fine-grained obstacles. In recent years, an increasing number of approaches \cite{ren2025safety}, \cite{wang2022geometrically} use flight corridors to characterize free space, improving the efficiency of interaction between motion planning algorithms and environmental perception. Nevertheless, flight corridors implicitly filter environmental topological information, leading to loss of solution space. 

Recent work has increasingly explored learning-based navigation methods for static environments. Zhang et al. \cite{zhang2025corb} use safe flight corridors as environmental representations, generating smooth B-spline trajectories through learning-based control point optimization. This approach achieves robust navigation in static cluttered environments and can avoid some low-speed objects. However, this approach is constrained by the speed of corridor generation, preventing rapid responses to high-speed obstacles. More recently, end-to-end approaches based on depth images \cite{lu2024you} and LiDAR point clouds \cite{xu2025flying} directly map perception inputs to control-related outputs, showing faster response times. However, due to the lack of high-frequency planning capability or temporal prediction ability, neither of the works is suitable for highly dynamic environments. In comparison, our method achieves reactive avoidance of moving obstacles by predicting a future action sequence at a high frequency.

\subsection{Dynamic Environment Navigation}
To address the need for obstacle avoidance in dynamic obstacle environments, researchers develop some dynamic obstacle detection algorithms to perceive and predict moving obstacles for safe trajectory planning. Lu et al. \cite{lu2024fapp} employ a clustering algorithm to segment dynamic obstacles from point clouds and utilize a Kalman filter to estimate the velocity of moving objects. Subsequently, an optimization algorithm is applied to generate the drone's trajectory. However, the point cloud-based dynamic perception methods are computationally intensive, and the accuracy significantly decreases when dynamic objects move at high speeds. Furthermore, solving model predictive control and polynomial optimization problems is highly time-consuming for complex scenarios. Unlike approaches relying on active perception, Liu et al. \cite{liu2023integrated} propose an integrated planning-and-control architecture. By combining high-frequency control responses with trajectory prediction capabilities, it enhances the system's reaction speed to dynamic obstacles, enabling avoidance of suddenly appearing moving objects.

Recently, significant progress has been made in learning-based dynamic obstacle avoidance. Xu et al. \cite{xu2025navrl} propose the NavRL framework, which estimates the states of dynamic obstacles and integrates a velocity-based safety shield to maintain safe flight in dynamic environments. However, due to limitations in object perception algorithms, real-time performance degrades significantly as the number of dynamic obstacles increases, making it challenging to maintain safety in highly dynamic settings. To enhance response ability, an increasing number of end-to-end approaches have been applied for obstacle avoidance in dynamic environments. Fan et al. \cite{fan2025flying} achieve dynamic obstacle perception by compressing multiple frames of previously scanned point clouds into a 2D obstacle map, then directly mapping the perceived information and state information into acceleration commands to control the quadrotor. Xu et al. \cite{xu2025flow} encode LiDAR point clouds into low-resolution, fixed-shape distance maps, combining them with multi-frame point stream information extracted by pre-trained NeuFlowV2 to form a perceptual representation of environmental changes. However, these approaches are typically developed with a strong emphasis on dynamic scenarios, causing a misalignment with static navigation tasks and consequently affecting their performance in static settings. In contrast, within our approach, the policy is only trained in static environments and enhances dynamic obstacle avoidance through high-frequency control responses and finite-horizon action-sequence prediction, while maintaining strong performance in static environments.

\section{Method}
\begin{figure*}[ht]
    \centering
    \includegraphics[width=0.95\textwidth]{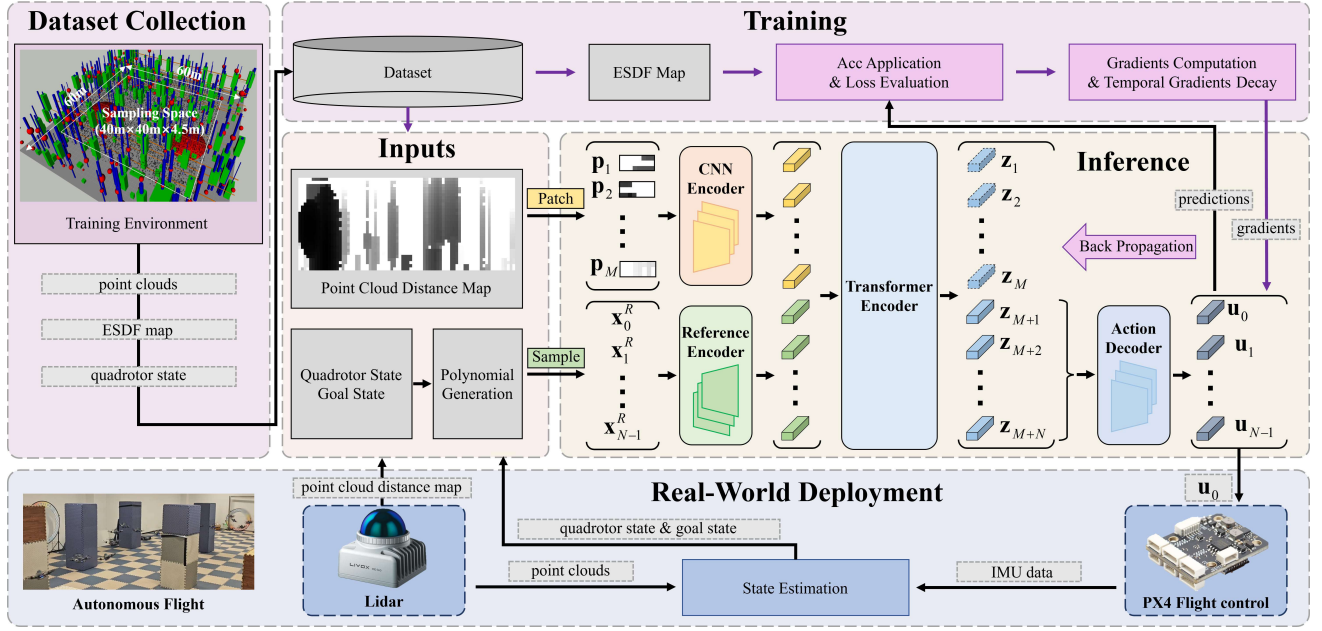}
    \caption{System architecture. Patches of the distance map and sampled states are encoded respectively and then concatenated into a token sequence to be input into the transformer encoder for predicting the acceleration sequence within a time horizon. ESDF map is obtained from the training dataset and is used only during training to evaluate the loss function and compute the gradients. The dataset is sampled from ten randomly generated training environments. In real-world deployment, the state of quadrotor is obtained using the state estimation method (e.g., Fast-LIO2\cite{xu2022fast}, and Point LIO\cite{he2023point}).}
    \label{fig:system}
\end{figure*}
The proposed TransNav architecture is illustrated in Fig. \ref{fig:system}, consisting of four components: point-cloud encoding, policy architecture, training strategy, and real-world deployment.

\subsection{Point Cloud Data Encoding}
\label{sec:point cloud encoding}
\begin{figure}[!t]
    \centering
    \includegraphics[width=\columnwidth]{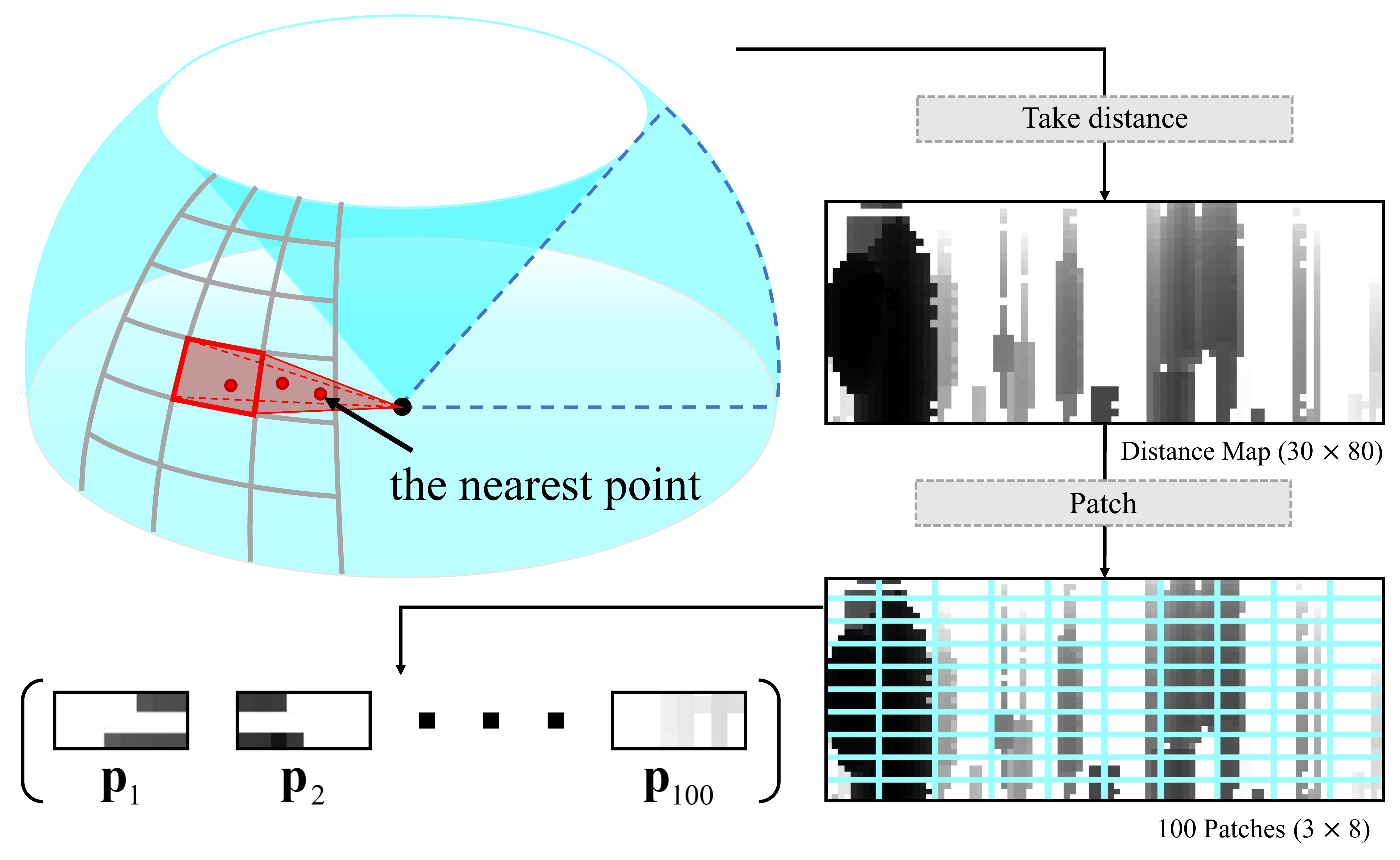}
    \caption{Illustration of patch-based LiDAR sensing representation. The distance map is composed of the distances from the nearest point of each partition, which are divided into 100 patches to be input into the subsequent transformer encoder for feature extraction.}
    \label{fig:point_cloud}
\end{figure}

To efficiently encode raw LiDAR measurements, we convert the point clouds into a structured distance map and further partition it into fixed-size local patches. As shown in Fig. \ref{fig:point_cloud}, the LiDAR sensing space is discretized according to angular resolution, and for each partition, only the distance of the nearest point to the quadrotor is retained. The Livox Mid-360 LiDAR has a field of view of $360^{\circ}\times59^{\circ}$. In our implementation, we choose a horizontal angular resolution of $4.5^{\circ}$ and a vertical angular resolution of $2^{\circ}$. The horizontal and vertical fields of view are divided into 80 and 30 partitions, respectively, yielding a total of 2400 partitions. Each partition stores a single depth value, resulting in a compact distance map of size $30\times 80$, analogous to a depth image.   

Then, to facilitate transformer-based processing while preserving spatial locality, the distance map is further divided into $M = 100$ fixed-size patches, each of dimension $3\times8$, maintaining the original aspect ratio. These patches are then used as the perception input to the policy network. Compared with grid maps or flight corridors, this representation reduces the cost of processing large raw point clouds, preserves the local geometric structure around obstacles, and provides a compact input for modeling long-range spatial relationships in later transformer layers.

\subsection{System Overview of Policy}
\label{sec:system}
This subsection defines the policy used in TransNav. The policy takes patch-based LiDAR perception and sampled reference-trajectory states as input, and outputs a finite-horizon action sequence for receding-horizon execution.

\textit{1) Observation Space}: As shown in Fig. \ref{fig:system}, the observation space $\mathbf{o}$ comprises two components: the perception parts derived from the LiDAR point clouds and the navigation parts derived from a reference trajectory. Both components are defined in the vehicle body frame. The perception input consists of the distance-map patches described in Sec. \ref{sec:point cloud encoding}, represented as a tensor of size $(M,3,8)$ with $M=100$. The navigation input is constructed from a reference quintic polynomial, which is uniquely determined by the initial state $(\mathbf{p}_0,\mathbf{v}_0,\mathbf{a}_0)$ and the target state $(\mathbf{p}_g,\mathbf{v}_g,\mathbf{a}_g)$. For each axis $\mu\in(x,y,z)$, the trajectory is given by:
$$
p(t)=c_0+c_1t+c_2t^2+c_3t^3+c_4t^4+c_5t^5, \eqno{(1)}
$$
with corresponding velocity obtained as
$$
v(t)=c_1+2c_2t+3c_3t^2+4c_4t^3+5c_5t^4. \eqno{(2)}
$$
The trajectory is sampled at regular intervals of 0.05 s over the prediction horizon to obtain $N=30$ reference states. Each reference state contains position and velocity, resulting in an $(N,6)$-dimensional navigation input.

\textit{2) Action Space}: The policy outputs a finite-horizon action sequence $\{\mathbf{u}_0,\mathbf{u}_1,...,\mathbf{u}_{N-1}\}$ corresponding to the reference state sequence. Each action $\mathbf{u}_k \in \mathbb{R}^3$ represents the desired acceleration in the vehicle body frame, and only the first action is executed at each control step in a receding-horizon manner. This design provides a short-term control tendency optimized over future states, while receding-horizon execution keeps the policy reactive to newly observed obstacles. At training time, the simple dynamics model directly treats the predicted action as acceleration input to propagate the state over the prediction horizon. During deployment, the predicted acceleration is converted into lower-level control commands through a thrust controller, which are then fed into the flight controller for execution. Compared with directly predicting thrust and bodyrates, using desired acceleration as the action improves the sim-to-real generalization capability of the policy.

\textit{3) Policy Representation}: The policy parameters are denoted by $\theta$, and the policy $\pi_\theta$ maps the observation defined above to a finite-horizon acceleration sequence through an encoder-only transformer network. First, the patch-based perception input and the sampled reference states are encoded separately. Each distance-map patch of size $3\times8$ is treated as a single-channel image and processed by a lightweight CNN. Specifically, the CNN contains two convolutional layers, whose structure [output channel, kernel size, stride, padding] is respectively $[32, 2, 1, 1]$, $[64, 2, 1, 0]$. The resulting feature map is flattened and projected to a 128-dimensional embedding through a fully connected layer. Therefore, the $M=100$ LiDAR patches are encoded as $M$ perception tokens. Each sampled reference state is projected to the same 128-dimensional feature space using a linear layer. The perception tokens and the reference-state tokens are then concatenated to a unified token sequence. This sequence is processed by a two-layer transformer encoder with two attention heads and a feed-forward dimension of 512. The transformer maps the input token sequence to latent tokens $\{\mathbf{z}_1,\mathbf{z}_2,...,\mathbf{z}_{M+N}\}$. Subsequently, the latent tokens associated with the reference states are processed by a two-layer MLP action decoder to generate the finite-horizon action sequence $\{\mathbf{u}_0,\mathbf{u}_{1},...,\mathbf{u}_{N-1}\}$.

We adopt an encoder-only transformer structure so that the entire action sequence can be generated in a single forward pass. Different from the GRU-based recurrent policy in \cite{zhang2025learning}, which aggregates temporal context through a hidden state and predicts a single action at each step, TransNav both processes the spatial LiDAR patch tokens and reference-state tokens. The self-attention mechanism directly models their interactions over the prediction horizon, enabling the policy to generate a predictive action sequence.

\subsection{Training Strategy}
\label{sec:training}
We develop an efficient unsupervised training strategy for the proposed finite-horizon action-sequence policy.

\textit{1) Policy Optimization}: We train the policy using an unsupervised strategy to directly minimize the trajectory-level cost over a finite prediction horizon. At each decision step, the policy takes the current observation as input and predicts a sequence of future actions in a single forward pass. The resulting future state sequence is then propagated through a lightweight point-mass dynamics model, allowing the policy to be optimized directly with respect to trajectory-level navigation costs.

Specifically, given the prediction horizon $N$, the policy outputs $\{\mathbf{u}_0,\mathbf{u}_{1},...,\mathbf{u}_{N-1}\}=\pi_\theta(\mathbf{o})$. At time step $k, k\in[0,N-1]$, the state comprises three-dimensional position and velocity, denoted as $\mathbf{x}_{k}\in\mathbb{R}^6$. This optimization problem can be formulated as
$$
\min_\theta L_{\theta}=\sum_{k=0}^{N-1}l_{k}=\sum_{k=0}^{N-1}l(\mathbf{x}_{k},\mathbf{u}_{k}), \eqno{(3a)}
$$
$$
 \text{s.t. }\{\mathbf{u}_0,\mathbf{u}_{1},...,\mathbf{u}_{N-1}\}=\pi_\theta(\mathbf{o}), \eqno{(3b)}
$$
where $l(\cdot)$ denotes the stage cost over the predicted horizon. The gradient for the above optimization problem can be expressed as follows:
$$
\frac{\mathrm{d}L_{\theta}}{\mathrm{d}\theta}=\sum_{k=0}^{N-1}\left[\frac{\partial l_{k}}{\partial\mathbf{x}_{k}}\frac{\mathrm{d}\mathbf{x}_{k}}{\mathrm{d}\theta}+\frac{\partial l_{k}}{\partial\mathbf{u}_{k}}\frac{\mathrm{d}\mathbf{u}_{k}}{\mathrm{d}\theta}\right]. \eqno{(4)}
$$
The chain propagation of gradients relies on the dynamics function. Inspired by \cite{zhang2025learning}, we use a point-mass model as the surrogate dynamics for state transitions to simplify computations. \cite{zhang2025learning} has demonstrated that the point-mass model accelerates gradient computation without degrading training performance. The dynamics function is denoted as $\mathbf{x}_{k+1}=f(\mathbf{x}_k,\mathbf{u}_k)$. Specifically, it can be written as
$$
\mathbf{p}_{k+1}=\mathbf{p}_k+\mathbf{v}_kdt+\frac{1}{2}\mathbf{a}_kdt^2,\quad\mathbf{v}_{k+1}=\mathbf{v}_k+\mathbf{a}_kdt, \eqno{(5)}
$$
where $\mathbf{p}_k \in \mathbb{R}^3$, $\mathbf{v}_k \in \mathbb{R}^3$, and $\mathbf{a}_k \in \mathbb{R}^3$ represent the position, velocity, and acceleration of the system, respectively, and $dt$ is the time interval. Applying the dynamical model to the chain rule gives the recursive relation as follows:
$$
\frac{\mathrm{d}\mathbf{x}_{k+1}}{\mathrm{d}\theta}=\frac{\partial\mathbf{x}_{k+1}}{\partial\mathbf{x}_{k}}\frac{\mathrm{d}\mathbf{x}_{k}}{\mathrm{d}\theta}+\frac{\partial\mathbf{x}_{k+1}}{\partial\mathbf{u}_{k}}\frac{\mathrm{d}\mathbf{u}_{k}}{\mathrm{d}\theta}. \eqno{(6)}
$$
Note that $\frac{\mathrm{d}\mathbf{x}_0}{\mathrm{d}\theta}=0$ since the initial state $\mathbf{x}_0$ is independent of the policy at each decision step.

\begin{figure}[!t]
    \centering
    \includegraphics[width=\columnwidth]{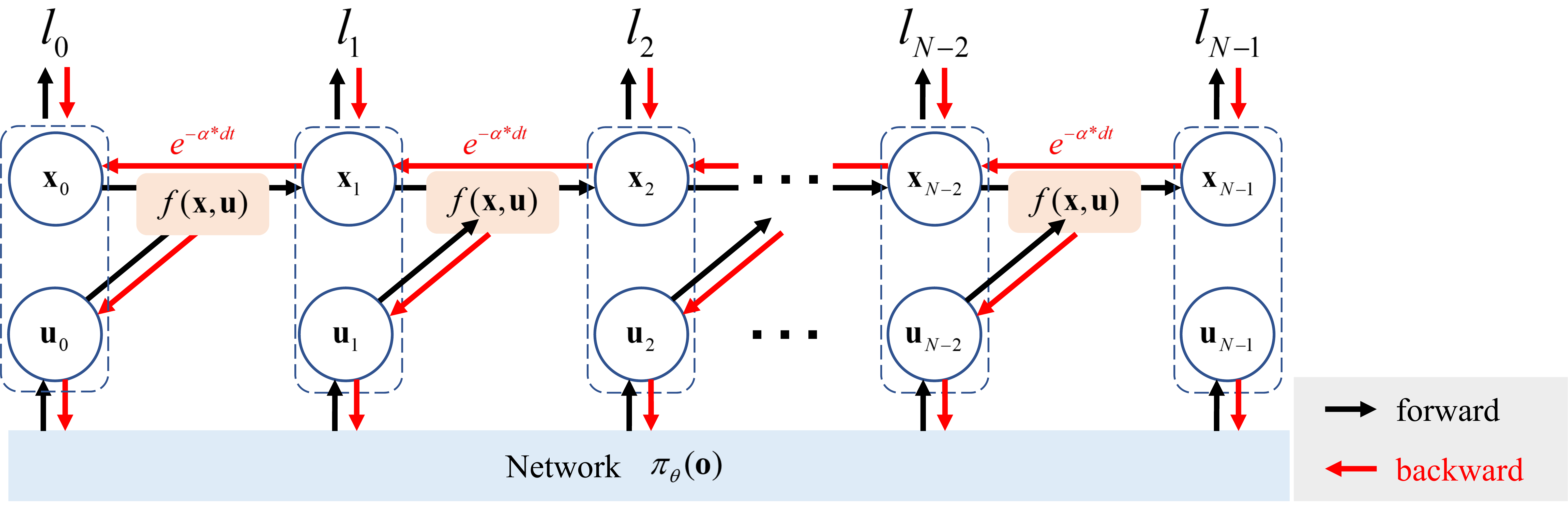}
    \caption{Illustration of backpropagation process. The output action sequence is applied to the simple point-mass model to obtain the state sequence prediction in future horizon. The complete gradient of the loss with respect to the network parameters is obtained directly through backpropagation, and the temporal gradient decay is used to alleviate gradient explosion during the backpropagation process.}
    \label{fig:backpropagation}
\end{figure}

Due to direct backpropagation across the entire time domain, gradients for future steps accumulate progressively during the backpropagation process, potentially leading to gradient explosion. To prevent the issue and suppress excessively premature avoidance responses to distant obstacles, we employ the exponential decay mechanism proposed in \cite{zhang2025learning} to suppress gradient accumulation. Specifically, the recursive gradient term with respect to the previous state is weighted by $e^{-\alpha dt}$, yielding
$$
\frac{\mathrm{d}\mathbf{x}_{k+1}}{\mathrm{d}\theta}=e^{-\alpha dt}\frac{\partial\mathbf{x}_{k+1}}{\partial\mathbf{x}_{k}}\frac{\mathrm{d}\mathbf{x}_{k}}{\mathrm{d}\theta}+\frac{\partial\mathbf{x}_{k+1}}{\partial\mathbf{u}_{k}}\frac{\mathrm{d}\mathbf{u}_{k}}{\mathrm{d}\theta}, \eqno{(7)}
$$
where $\alpha$ is the decay coefficient, which is set to 0.9 in our implementation. The complete gradient flow is illustrated in Fig. \ref{fig:backpropagation}. After computing the gradient, we employ the Adam optimizer to optimize the policy. 

The proposed training strategy is motivated to efficiently train a policy that outputs a finite-horizon action sequence. For imitation learning, it is computationally expensive to obtain large-scale expert demonstrations because expert generation typically depends on front-end search and back-end trajectory optimization \cite{liu2023integrated}. Moreover, regression to expert actions does not necessarily align with the trajectory-level costs that ultimately determine navigation quality \cite{lu2024you}. In contrast, our strategy evaluates trajectory-level navigation costs directly on the predicted rollout, avoiding the need for expert labels. Since the policy outputs an entire action sequence at once, applying critic-based RL to our policy is nontrivial. The resulting high-dimensional action space can lead to unstable training and slow convergence.

\textit{2) Loss Function Design}: The loss function consists of three components: a guidance loss $L_g$, a safety loss $L_s$, and a smoothness loss composed of $L_a$ and $L_j$, yielding
$$
L=\lambda_gL_g+\lambda_sL_s+\lambda_aL_a+\lambda_jL_j, \eqno{(8)}
$$
where $\lambda_g$, $\lambda_s$, $\lambda_a$, and $\lambda_j$ are the respective weights for each loss component. Specifically, we choose $\lambda_g=1.0$, $\lambda_s=1.5$, $\lambda_a=0.005$, and $\lambda_j=0.0005$ in our implementation.

Guidance loss $L_g$ contains an integral term that penalizes deviation from the desired velocity and a terminal term that penalizes deviation from the goal state:
$$
L_g=\sum_{k=1}^N\mathrm{Smooth~L}1(\parallel\mathbf{v}_k^{set}-\mathbf{v}_k\parallel,0)dt+0.5L_{term}, \eqno{(9a)}
$$
$$
L_{term}= \mathrm{Smooth~L}1(\parallel\mathbf{x}_N-\mathbf{x}_{goal}\parallel,0), \eqno{(9b)}
$$
where $\mathbf{v}_k \in \mathbb{R}^3$ is the velocity, $\mathbf{v}_k^{set} \in \mathbb{R}^3$ is the desired velocity, pointing from the current position toward the goal, with prescribed magnitude 7 m/s, $\mathbf{x}_N \in \mathbb{R}^6$ is the terminal state of the trajectory, $\mathbf{x}_{goal} \in \mathbb{R}^6$ is the goal state, and the state includes position and velocity.

Safety loss $L_s$ is defined as the integral of the safety function over the predicted trajectory:
$$
L_s=\sum_{k=1}^N\text{safe}(\mathbf{x}_k)dt. \eqno{(10)}
$$
For each predicted state, the safety function is defined as
$$
\text{safe}(\mathbf{x}_k)=v_k^o\max(d_0-d_k,0)+2e^{(-(d_k-d_0)/\beta)}, \eqno{(11)}
$$
where $d_k$ is the distance to the nearest obstacle, $v_k^o$ is the velocity component toward that obstacle, and $d_0$ and $\beta$ are the predefined safety-margin and scaling factor, respectively. 
Safety loss accounts for both immediate collision risk and the potential risk associated with approaching the nearest obstacle. The velocity $v_k^o$ and distance $d_k$ along with their gradients are derived from the ESDF map, which is the privileged information accessible only during training.

Smoothness loss is defined as a penalty for acceleration and jerk of the quadrotor:
$$
L_a=\sum_{k=1}^N\parallel a_k\parallel^2dt, L_j=\sum_{k=1}^N \bigg\| \frac{a_{k+1}-a_k}{dt}\bigg\|^2dt. \eqno{(12)}
$$
Smoothness loss ensures dynamically feasible trajectories by limiting the magnitude of acceleration and jerk.

\textit{3) Training Setup and Implementation}: The training dataset is collected from ten randomly generated static environments. As shown in Fig. \ref{fig:system}, each environment covers a volume of $(60\ \mathrm{m}\times 60\ \mathrm{m}\times 15\ \mathrm{m})$ with a sampling space $(40\ \mathrm{m}\times40\ \mathrm{m}\times4.5\ \mathrm{m})$. Each environment contains four types of obstacles: cylinders, cubes, spheres, and thin horizontal cylinders, with 100 obstacles for each type. We randomly sample the state of quadrotor in the collision-free sampling space. The goal position is sampled along a random direction with a distance randomly selected from 1 to 10 m. The goal velocity is randomly sampled from 0 to 7 m/s along the start-to-goal direction. The goal acceleration is set to zero. For each valid sample, the point clouds, quadrotor state, goal state, and ESDF map are stored, where the ESDF is only used for training loss evaluation and gradient computation. We collect 10 K samples from each environment and train the policy for 50 epochs with a batch size of 16, converging within two hours. The learning rate is set to $1\times 10^{-4}$, and the cosine annealing mechanism is employed to enhance the training efficiency.

\section{EXPERIMENTS}
In this section, we conduct a series of experiments to validate the performance of the proposed method in both simulated and real-world environments. All simulation experiments and policy training are performed on a desktop computer equipped with an Intel i9-12900K CPU and an NVIDIA GeForce RTX4060 GPU.

\begin{figure*}[!t]
    \centering
    \includegraphics[width=0.95\textwidth]{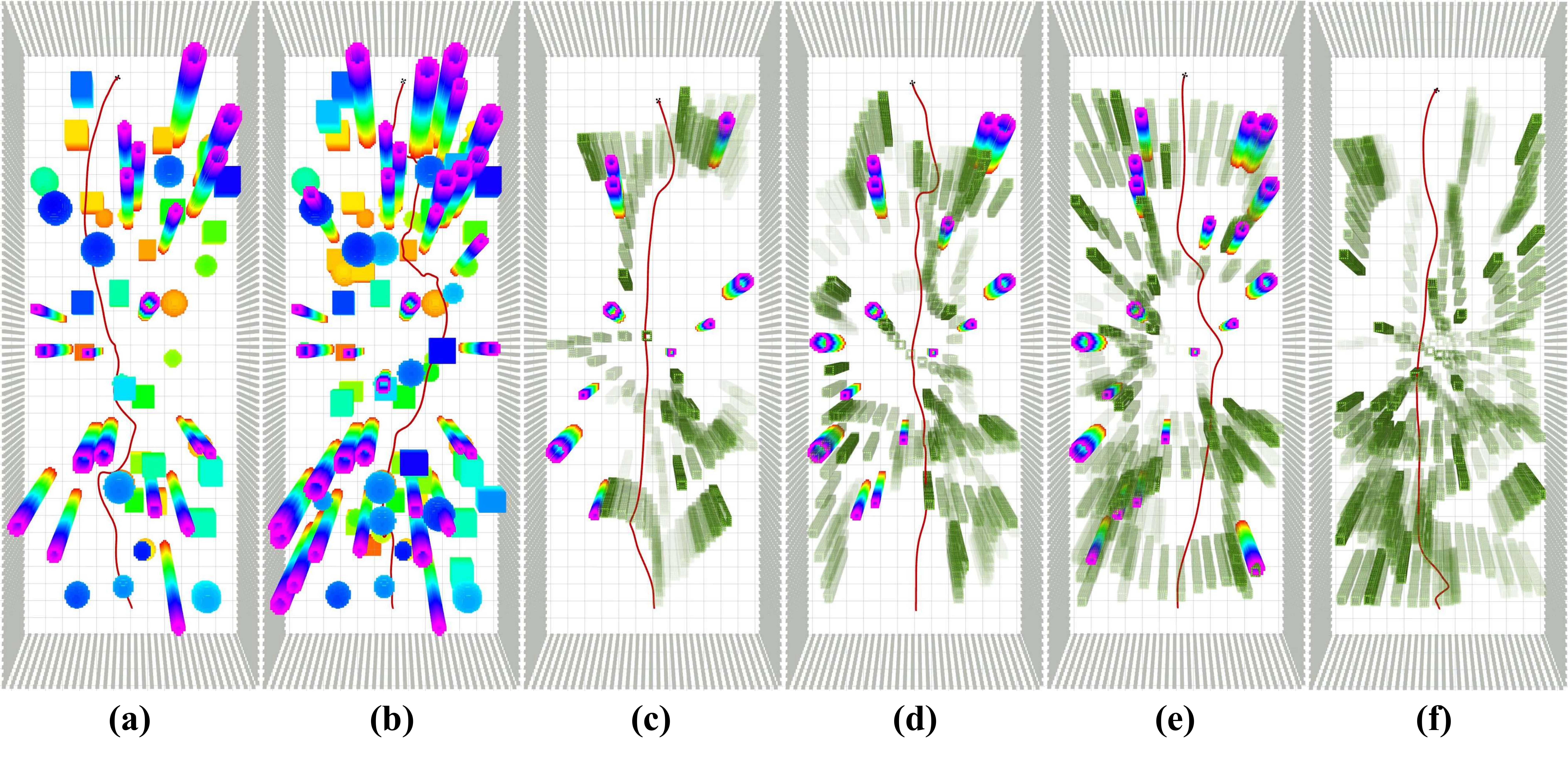}
    \caption{Visualization of flight slices in simulation. The scenarios labeled (a) to (f) correspond to scenarios 1 to 6 in table \ref{tab:simulation_results}, respectively.}
    \label{fig:simulation}
\end{figure*}

\begin{table}[!t]
  \centering
  \caption{SIMULATED COMPARISON IN DIFFERENT TYPES OF ENVIRONMENTS.}
  \begin{tabular}{>{\centering\arraybackslash}p{1.5cm}  
    >{\centering\arraybackslash}p{0.9cm}
    >{\centering\arraybackslash}p{0.9cm}
    >{\centering\arraybackslash}p{0.9cm}
    >{\centering\arraybackslash}p{0.9cm}
    >{\centering\arraybackslash}p{0.9cm}
    }
    \toprule
    \textbf{Scenario} & \textbf{Method} & $\eta(\%)$ & $v_a(m/s)$ & $t_p(ms)$ & $R_l$ \\
    \midrule
    \multirow{5}{*}{\makecell{60 static}}
    & Ours & \cellcolor{gray!25}100 & \cellcolor{gray!10}2.71 & \cellcolor{gray!25}6.72  & \cellcolor{gray!10}1.12 \\
    & \cite{zhou2020ego}  & \cellcolor{gray!10}85  & \cellcolor{gray!25}3.12 &  11.73  & \cellcolor{gray!25}1.05 \\
    & \cite{lu2024fapp} & 80  & 2.63 & 9.59  & \cellcolor{gray!25}1.05 \\
    & \cite{xu2025navrl} & \cellcolor{gray!25}100 & 0.90 & 23.14 & \cellcolor{gray!25}1.05 \\
    & \cite{xu2025flow}  & 60  & 2.46 & \cellcolor{gray!10}8.43  & 2.07 \\
    \midrule
    \multirow{5}{*}{\makecell{105 static}}
    & Ours & \cellcolor{gray!25}100 & 1.97 & \cellcolor{gray!25}7.01  & 1.19 \\
    & \cite{zhou2020ego}  & \cellcolor{gray!10}80  & \cellcolor{gray!25}2.89 &  16.91     & \cellcolor{gray!25}1.06 \\
    & \cite{lu2024fapp} & 40  & 2.04 & 37.04 & \cellcolor{gray!10}1.17 \\
    & \cite{xu2025navrl} & \cellcolor{gray!10}80  & 0.95 & 22.87 & 1.22 \\
    & \cite{xu2025flow}  & 10  & \cellcolor{gray!10}2.28 & \cellcolor{gray!10}7.70  & 2.33 \\
    \midrule
    \multirow{5}{*}{\makecell{10 dynamic\\ and 10 static}}
    & Ours & \cellcolor{gray!25}80  & \cellcolor{gray!25}3.77 & \cellcolor{gray!25}6.87  & \cellcolor{gray!25}1.05 \\
    & \cite{zhou2020ego}  & 5   & \cellcolor{gray!10}3.31 & \cellcolor{gray!10}10.06   & \cellcolor{gray!10}1.08 \\
    & \cite{lu2024fapp} & \cellcolor{gray!10}70  & 2.44 & 15.99 & 1.14 \\
    & \cite{xu2025navrl} & 40  & 0.88 & 24.44 & 1.06 \\
    & \cite{xu2025flow}  & \cellcolor{gray!25}80  & 3.05 & 11.11 & 1.11 \\
    \midrule
    \multirow{5}{*}{\makecell{15 dynamic\\ and 15 static}}
    & Ours & \cellcolor{gray!25}60  & \cellcolor{gray!25}3.08 & \cellcolor{gray!25}7.06  & \cellcolor{gray!10}1.13 \\
    & \cite{zhou2020ego}  & N/A & N/A  & N/A   & N/A  \\
    & \cite{lu2024fapp} & \cellcolor{gray!10}35  & 2.28 & 18.99 & 1.21 \\
    & \cite{xu2025navrl} & 20  & 0.86 & 24.86 & \cellcolor{gray!25}1.10 \\
    & \cite{xu2025flow}  & \cellcolor{gray!25}60  & \cellcolor{gray!10}3.02 & \cellcolor{gray!10}11.22 & 1.19 \\
    \midrule
    \multirow{5}{*}{\makecell{20 dynamic\\ and 20 static}}
    & Ours & \cellcolor{gray!25}40  & \cellcolor{gray!25}3.07 & \cellcolor{gray!25}6.42  & \cellcolor{gray!10}1.07 \\
    & \cite{zhou2020ego}  & N/A & N/A  & N/A   & N/A  \\
    & \cite{lu2024fapp} & 15  & 2.20 & 24.72 & 1.10 \\
    & \cite{xu2025navrl} & 15  & 0.86 & 19.60 & \cellcolor{gray!25}1.06 \\
    & \cite{xu2025flow}  & \cellcolor{gray!10}30  & \cellcolor{gray!10}2.77 & \cellcolor{gray!10}10.70 & 1.32 \\
    \midrule
    \multirow{5}{*}{\makecell{25 dynamic}}
    & Ours & \cellcolor{gray!10}35  & \cellcolor{gray!25}3.22 & \cellcolor{gray!25}6.13  & 1.13 \\
    & \cite{zhou2020ego}  & N/A & N/A  & N/A   & N/A  \\
    & \cite{lu2024fapp} & N/A & N/A  & N/A   & N/A  \\
    & \cite{xu2025navrl} & 5   & 0.91 & 24.30 & \cellcolor{gray!25}1.11 \\
    & \cite{xu2025flow}  & \cellcolor{gray!25}40  & \cellcolor{gray!10}3.11 & \cellcolor{gray!10}10.99 & \cellcolor{gray!10}1.12 \\
    \bottomrule
  \end{tabular}
  \label{tab:simulation_results}
\end{table}

\subsection{Simulation Experiments}
We compare the proposed method with four representative baselines: the optimization-based static planner EGO-Planner \cite{zhou2020ego}, the optimization-based dynamic planner FAPP \cite{lu2024fapp}, the hierarchical learning-based method NavRL \cite{xu2025navrl}, and the end-to-end learning-based method P2M \cite{xu2025flow}. To ensure fair comparison, all methods are configured with identical sensing ranges and perception update frequencies. For optimization-based methods \cite{zhou2020ego}, \cite{lu2024fapp}, the velocity and acceleration limits are set to 3.5 m/s and 10 m/s$^2$, respectively. \cite{xu2025navrl} constrains its velocity output to [0.5,1.5] m/s. The reference velocity and acceleration of \cite{xu2025flow} are set to 5 m/s and 10 m/s$^2$. For the proposed method, the goal velocity is bounded by 7 m/s and the goal acceleration is set to zero.

To comprehensively evaluate the performance, two simulation scenarios are designed for benchmarking: 1) dense static environment that contains three types of static obstacles: cylinders, cubes, and spheres; 2) highly dynamic environment that contains static cylindrical obstacles and dynamic cubic obstacles executing variable-speed curved motions with maximum velocity up to 5 m/s. Each environment spans an obstacle area of 25 m$\times$10 m. The simulation environments are shown in Fig. \ref{fig:simulation}. For each scenario, we count the success rate $\eta$ from 20 trials and other performance metrics from 5 successful flights, including average flight speed $v_a$, latency $t_p$ from perception to action planning, and path efficiency $R_l$, defined as the ratio between the realized path length and the straight-line distance from start to goal. When there are fewer than five successful trials, all available successful flights are used to calculate these averages.

\begin{figure*}[ht]
    \centering
    \includegraphics[width=0.95\textwidth]{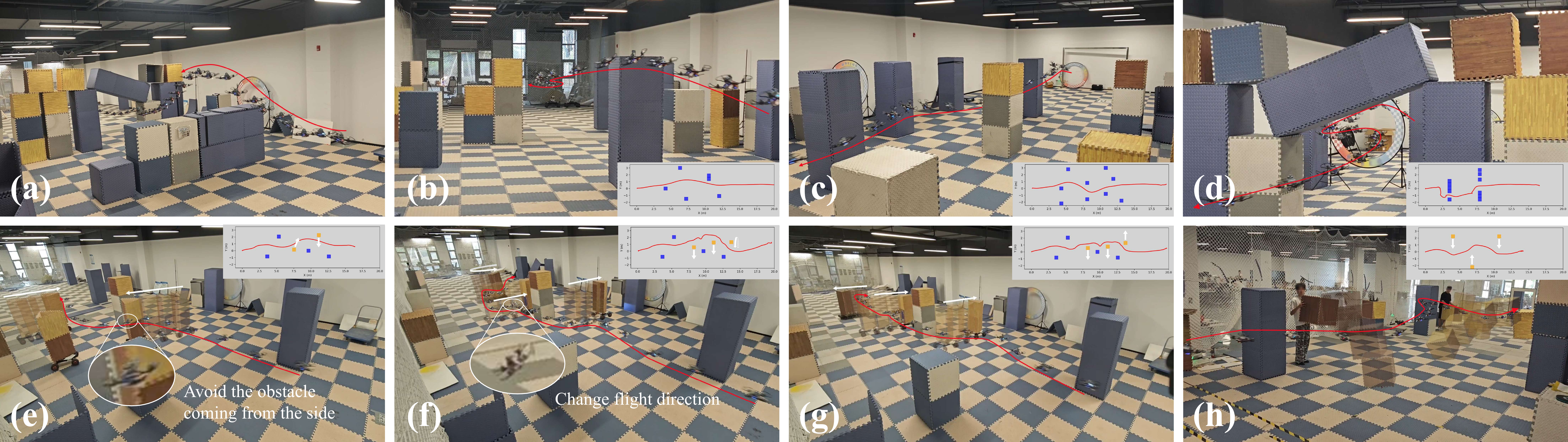}
    \caption{Real-world experiments. (a) Highlight of a vertical avoidance capability. (b)-(c) Agile flight among static obstacles. (d) Flight through a gate. (e)-(g) Navigation among static and dynamic obstacles. (h) Avoiding fast-moving obstacles. White curves represent the movement paths of dynamic obstacles.}
    \label{fig:real-world}
\end{figure*}
The results in Table \ref{tab:simulation_results} show that our method performs particularly well in dense static environments and remains competitive in scenarios involving moving obstacles. In the two static settings, TransNav achieves the highest success rate, indicating that the patch-based LiDAR representation preserves useful geometric information for cluttered navigation. Although Ego-Planner \cite{zhou2020ego} achieves a higher average speed, TransNav adaptively reduces speed in more complex environments to maintain a high success rate. In dynamic scenarios, TransNav maintains competitive success rates, suggesting that finite-horizon action-sequence prediction and efficient transformer-based policy inference improve responses to environmental changes. Failures in highly dynamic scenarios mainly occur when fast-approaching obstacles leave insufficient time for the quadrotor to change its velocity and move away, since the dynamics prevent instantaneous evasive motion. In the scenario that only contains dynamic obstacles, our method is slightly outperformed by \cite{xu2025flow} in success rate due to the lack of prediction for dynamic obstacles. However, the performance of \cite{xu2025flow} degrades substantially in dense static environments. Overall, the simulation results indicate that TransNav learns a more general navigation policy across different obstacle configurations. Thanks to the lightweight network structure, our method achieves the lowest latency among all benchmarks.

\subsection{Ablation Studies}
\begin{table}[!t]
  \centering
  \caption{ABLATION STUDY}
  \begin{tabular}{>{\centering\arraybackslash}p{1.5cm}  
    >{\centering\arraybackslash}p{1.8cm}
    >{\centering\arraybackslash}p{0.9cm}
    >{\centering\arraybackslash}p{0.9cm}
    >{\centering\arraybackslash}p{1.2cm}
    }
    \toprule
    \textbf{Scenario} & \textbf{Ablation} & $\eta(\%)$ & $v_a(m/s)$ & Parameters  \\
    \midrule
    \multirow{6}{*}{\makecell{60 static}}
    & Ours & \cellcolor{gray!25}100 & \cellcolor{gray!25}2.71 & \cellcolor{gray!25}0.6M  \\
    & w/o decay  & \cellcolor{gray!25}100  & 1.07 & \cellcolor{gray!25}0.6M \\
    & w/o patch  & \cellcolor{gray!10}50  & \cellcolor{gray!10}1.20 & \cellcolor{gray!10}1.7M  \\
    & w/o sequence & N/A & N/A & \cellcolor{gray!25}0.6M \\
    & encoders & N/A & N/A & $\geq$3.7M \\
    & PPO & N/A & N/A & \cellcolor{gray!25}0.6M \\
    \midrule
    \multirow{6}{*}{\makecell{10 dynamic\\ and 10 static}}
    & Ours & \cellcolor{gray!25}80 & \cellcolor{gray!10}3.77 & \cellcolor{gray!25}0.6M  \\
    & w/o decay  & 40  & 1.97 & \cellcolor{gray!25}0.6M \\
    & w/o patch  & \cellcolor{gray!10}45  & 2.37 & \cellcolor{gray!10}1.7M \\
    & w/o sequence & 20 & \cellcolor{gray!25}6.51 & \cellcolor{gray!25}0.6M \\
    & encoders & N/A & N/A & $\geq$3.7M \\
    & PPO & N/A & N/A & \cellcolor{gray!25}0.6M \\
    \bottomrule
  \end{tabular}
  \label{tab:ablation_results}
\end{table} 
We perform ablation studies to examine the key components of the proposed method. Specifically, the “w/o decay” variant sets the temporal decay coefficient $\alpha$ in Eq. (7) to 0, the “w/o patch” variant processes the entire distance map with a CNN without patch partitioning, and the “w/o sequence” variant predicts a single acceleration command instead of an action sequence. For encoder ablation, we replace the transformer-based encoder with ResNet-18, a PointNet-style encoder, and UNet. The resulting perception feature is concatenated with the encoded reference states and fed into an MLP for action-sequence prediction. For training-strategy comparison, we train the same policy architecture using PPO. The evaluation is performed in the scenarios corresponding to (a) and (c) outlined in Fig. \ref{fig:simulation}, and the results are shown in Table \ref{tab:ablation_results}. 

Without temporal gradient decay, the success rate in the mixed scenario drops from 80\% to 40\%, and the average speed decreases significantly in all scenarios. The performance degradation is caused by the policy’s premature response to distant obstacles, resulting in overly conservative behavior. Without patch-based perception, both the success rate and average speed across all scenarios exhibit a substantial decline. These results indicate that the patch-based representation helps preserve compact but structured geometric information for navigating narrow spaces. When replacing sequence prediction with single-step output, we notice that the average speed rises from 3.77 m/s to 6.51 m/s in the mixed scenario, but the policy is virtually incapable of obstacle avoidance. This suggests that single-step prediction encourages aggressive goal-approaching behavior but cannot account for future obstacle interactions. The encoder comparisons show that replacing the fusion module with ResNet-18, PointNet-style, or UNet leads to failed navigation in the tested scenarios. The reason lies in that these alternatives are less effective in capturing the spatial-temporal relationship between the reference-state sequence and the surrounding obstacle distribution. In addition, PPO fails to train the same action-sequence policy stably, mainly due to the high-dimensional action space introduced by sequence prediction. Overall, these results demonstrate that the proposed performance benefits from the joint effect of the self-attention-based network and the trajectory-level training strategy, rather than from any single component.

\subsection{Real-World Experiments}
We develop a quadrotor with a 200 mm wheelbase and a total mass of 755 $g$ in Fig. \ref{fig:uav}. The platform is equipped with an Orange Pi 5 Pro onboard computer that features a Rockchip RK3588S processor. For environmental perception, a Livox Mid-360 LiDAR is used to provide raw point clouds. Although the intrinsic scan frequency of the LiDAR is 10 Hz, we incrementally update the representation using newly received partial point-cloud measurements, which enables the perception input to be refreshed at 100 Hz. With RKNN deployment, the policy inference time is approximately 5 ms, enabling control-command generation at 100 Hz on the onboard computer. The desired yaw is generated according to the horizontal velocity direction. The controller only executes the first command in a receding-horizon manner, which avoids open-loop accumulation of errors under the simplified dynamics model and mitigates the sim-to-real gap.

\textit{1) Static Environments}: To test the vertical avoidance capability, we design a challenging scenario by using square boxes to construct a low wall and a door frame. Neither of these obstacles is included in the dataset. As shown in Fig. \ref{fig:real-world}(a), the quadrotor can adaptively navigate through obstacles of varying heights without collisions. To test the high-speed flight performance of our approach, we retain a small number of obstacles in the test environment and increase the agility of the policy inputs. The flight trajectory is illustrated in Fig. \ref{fig:real-world}(b), with a maximum speed of 6.05 m/s and an average speed of 4.30 m/s. Keeping other conditions constant, we increase the number of static obstacles to raise the difficulty. As shown in Fig. \ref{fig:real-world}(c), the quadrotor can autonomously reduce flight speed to accommodate the increased obstacle density, achieving a maximum speed of 4.45 m/s. In 10 tests, our method succeeds 10 times. We further evaluate the narrow-space navigation using a circular gate with a diameter of 1 m. As shown in Fig. \ref{fig:real-world}(d), the quadrotor still maintains safe flight.

\textit{2) Dynamic Environments}: 
To evaluate the navigation capabilities of the policy in complex dynamic environments, we design a scenario that contains both static and dynamic obstacles. As shown in Fig. \ref{fig:real-world}(e), the scenario includes two moving obstacles, both of which perform a uniformly accelerated motion at a maximum speed of 2 m/s. The quadrotor effectively navigates through the obstacle area while maintaining a safe distance from all obstacles. Furthermore, due to the $360^\circ$ horizontal field of view, the quadrotor can accelerate away from the obstacle that rapidly approaches from the side. In 10 tests, our method succeeds 9 times.

We further increase the challenge by adding an extra dynamic obstacle. Building on the scenario shown in Fig. \ref{fig:real-world}(e), we introduce a third moving obstacle that is manually controlled to continuously block the quadrotor's flight path. As shown in Fig. \ref{fig:real-world}(f)-(g), the policy successfully maneuvers around the third obstacle. Furthermore, when the risk of collision increases due to the rapid movement of dynamic obstacles, the policy does not rigidly adhere to the original flight path but instead adaptively switches direction to flexibly avoid obstacles. To evaluate the policy’s response to fast-moving objects, we constantly throw boxes to interfere with the quadrotor. As shown in Fig. \ref{fig:real-world}(h), the quadrotor successfully reacts to multiple fast-moving objects, highlighting the combined effect of high-frequency replanning and finite-horizon action prediction. Compared with a depth camera, LiDAR provides a wider sensing range for dynamic obstacle avoidance. However, the low update frequency and redundant sensing data limit the flight speed of quadrotors.

\section{CONCLUSIONS}
This paper considered quadrotor obstacle avoidance in environments that contained both dense static obstacles and fast-moving objects. We proposed TransNav, a transformer-based end-to-end navigation method that integrated patch-based point cloud encoding with global self-attention to jointly achieve perception and motion. By predicting future action sequences at a high frequency, the policy enabled reactive obstacle avoidance without explicit obstacle tracking or motion prediction. Moreover, an efficient unsupervised training strategy was designed, allowing trajectory-level costs to be directly optimized through backpropagation. Finally, extensive simulation and real-world experiments demonstrated that our method provided a general performance for obstacle avoidance across static and dynamic settings within a single policy. Future work will extend the proposed framework to more challenging multi-agent and extreme-speed scenarios.
\addtolength{\textheight}{-12cm}   


 
\bibliographystyle{IEEEtran}
\bibliography{main}

\newpage

\vfill

\end{document}